\pdfoutput=1

\documentclass[11pt]{article}

\usepackage[final]{acl}

\usepackage{times}
\usepackage{latexsym}

\usepackage[T1]{fontenc}

\usepackage[utf8]{inputenc}

\usepackage{microtype}

\usepackage{inconsolata}

\usepackage{graphicx}

\usepackage{url}
\usepackage{float}                 
\usepackage{booktabs}              
\usepackage{amsmath}               
\usepackage[ruled]{algorithm2e}    
\usepackage{array}                 
\usepackage[normalem]{ulem}        
\useunder{\uline}{\ul}{}

\title{SELT: Self-Evaluation Tree Search for LLMs with Task Decomposition}

\author{
  \textbf{Mengsong Wu\textsuperscript{1,2}},
  \textbf{Di Zhang\textsuperscript{2}},
  \textbf{Yuqiang Li\textsuperscript{2}},
\\
  \textbf{Dongzhan Zhou\textsuperscript{2}},
  \textbf{Wenliang Chen\thanks{Corresponding Author.}\textsuperscript{1}}
\\
\\
  \textsuperscript{1}Soochow University,
  \textsuperscript{2}Shanghai Artificial Intelligence Laboratory
\\
  \small{ \texttt{radi.cat@qq.com, \{zhangdi,liyuqiang,zhoudongzhan\}@pjlab.org.cn, wlchen@suda.edu.cn}}
}

\begin{document}
\maketitle
\begin{abstract}

While Large Language Models (LLMs) have achieved remarkable success in a wide range of applications, their performance often degrades in complex reasoning tasks.
In this work, we introduce \textbf{SELT} (\textbf{S}elf-\textbf{E}valuation \textbf{L}LM \textbf{T}ree Search), a novel framework that leverages a modified Monte Carlo Tree Search (MCTS) to enhance LLM reasoning without relying on external reward models.
By redefining the Upper Confidence Bound scoring to align with intrinsic self-evaluation capabilities of LLMs and decomposing the inference process into atomic subtasks augmented with semantic clustering at each node, SELT effectively balances exploration and exploitation, reduces redundant reasoning paths, and mitigates hallucination.
We validate our approach on challenging benchmarks, including the knowledge-based MMLU and the Tool Learning dataset Seal-Tools, where SELT achieves significant improvements in answer accuracy and reasoning robustness compared to baseline methods.
Notably, our framework operates without task-specific fine-tuning, demonstrating strong generalizability across diverse reasoning tasks.
Relevant results and code are available at \url{https://github.com/fairyshine/SELT}.

\end{abstract}

\section{Introduction}

Large Language Models (LLMs) have demonstrated remarkable capabilities in compressing and memorizing vast amounts of knowledge \cite{Compressor, KnowledgeInjection}. 
However, LLMs exhibit significant weaknesses in complex reasoning tasks, often yielding inconsistent answers \cite{LogicBench}.
To augment reasoning abilities of LLMs, researchers have explored various approaches, such as In-Context Learning (ICL, \citealp{ICL}), Chain-of-Thought (CoT, \citealp{CoT}) prompting, and Reinforcement Learning (RL) based fine-tuning.
Despite successes of these methods, they often rely heavily on handcrafted reasoning templates or demand costly reward models which are extensively fine-tuned on domain-specific data.
Better Test-Time Scaling methods are still in need.

Recently, Monte Carlo Tree Search (MCTS, \citealp{MCTS_survey}) has emerged as a promising technique to enhance LLM inference by systematically investigating potential reasoning paths.
However, current methods \cite{LLM_MCTS, RAP} heavily rely on fine-tuning external reward models to evaluate intermediate reasoning steps, limiting their applications to domains with abundant task-specific data, such as mathematics.
It introduces additional training overhead and potential bias from imperfect reward signals.

\begin{figure}[htb]
  \centering
  \includegraphics[width=0.9\linewidth]{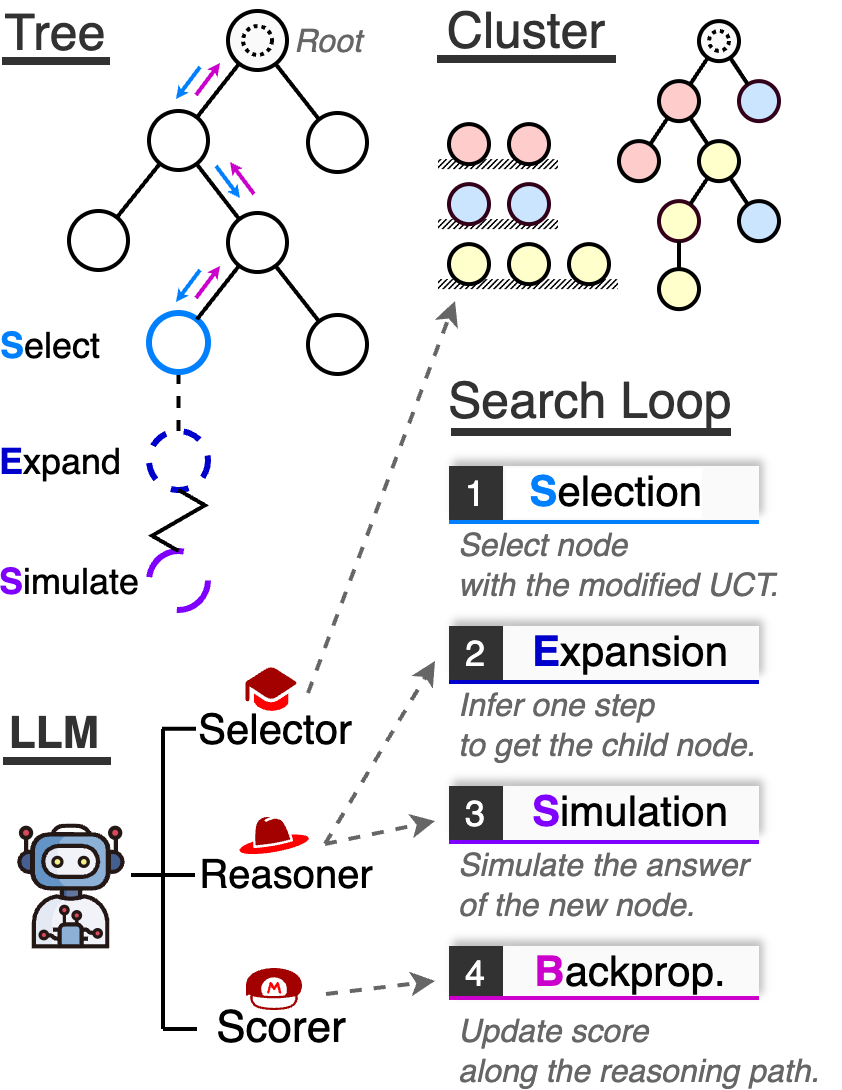}
  \caption{Tree Search with the Foundation Language Model and Clustering Strategy, without Fine-tuning Reward Model.}
  \label{fig:intro}
\end{figure}

In this work, we propose \textbf{SELT} (\textbf{S}elf-\textbf{E}valuate \textbf{L}LM \textbf{T}ree Search), a novel MCTS framework for LLMs designed to overcome these challenges through two key innovations.
First, we modify the raw MCTS algorithm by redefining two parts of the Upper Confidence Bound for Trees (UCT) scoring mechanism to align with LLM intrinsic self-evaluation capabilities, eliminating the need for external reward models.
This new scoring mechanism enables more balanced exploration and exploitation of reasoning paths.
Second, we decompose the inference process into multiple atomic LLM tasks, allowing the LLM to complete smaller, manageable subtasks.
To dynamically group semantically equivalent solutions, we introduce a semantic clustering mechanism at each tree node, thereby reducing redundancy and enabling the selection of higher-quality representative answers. 
With those representative answers, the LLM can self-evaluate the new answer better like in Figure \ref{fig:intro} without needing other reward models.
This approach not only enhances search efficiency but also mitigates the "hallucination trap" while preserving diversity in reasoning pathways.

We evaluate SELT on two challenging benchmarks: 
the knowledge-based question answering dataset \textbf{MMLU} \cite{MMLU}, which requires multi-step reasoning based on domain knowledge; and the Tool Learning dataset \textbf{Seal-Tools} \cite{Seal-Tools}, which involves dynamic interaction with external tools.
Experimental results demonstrate that SELT outperforms baseline method, including CoT and standard MCTS, achieving significant improvements in answer accuracy and reasoning robustness. 
Both innovations proposed are effective to enhance the LLM ability of answering complex questions.
Notably, our framework requires no task-specific fine-tuning, underscoring its generalizability across various reasoning tasks.

Our contributions are threefold: \\
\textbf{1)} \textbf{Self-Evaluation-Based Search}: We propose a self-evaluation-driven adaptation of MCTS for LLMs, eliminating the dependence on external reward models and allowing LLMs to guide the tree search intrinsically. \\ 
\textbf{2)} \textbf{LLM Inference Decomposition}: We simplify raw reasoning tasks into atomic subtasks, improving overall performance. In order to simplify the obtained reasoning path, we apply semantic-aware clustering to identify and prioritize diverse, high-quality solutions. \\
\textbf{3)} \textbf{Experimental Validation}: Extensive experiments demonstrate that SELT outperforms baseline methods in mathematical, commonsense, and procedural reasoning tasks, highlighting its generalizability without task-specific fine-tuning.

\section{Preliminary}

\subsection{List of Symbols}

To better illustrate the tree search algorithm for the LLM inference, the symbols which may present in the pseudo code are listed in Table \ref{table:symbols}.
\begin{table}[htb]
\centering
\resizebox{0.95\linewidth}{!}{
\begin{tabular}{>{\centering\arraybackslash}p{1cm}p{0.2cm}p{5cm}}
\toprule
\textbf{Symbol} & & \multicolumn{1}{c}{\textbf{Description}} \\
\midrule
$v$ & & One node in the reasoning trees which contain its structure information and the state. \\
$v_r$ & & Root node in the reasoning tree. \\
$s$ & & State of the specifc node which contains information about the LLM inference. \\
$\Delta$ & & Reward score given by the LLM. \\
$C_\beta$ & & Constant controlling the bayes averaging.\\
$C_p$ & & The exploration constant of MCTS, usually set to $\sqrt{2}$.\\
$Q(v)$ & & Total rewards of the node $v$.\\
$N(v)$ & & Visit times of the node $v$.\\
$r(s)$ & & The reward calculation of the state $s$.\\
\bottomrule
\end{tabular}
}
\caption{Relevant Symbols appeared in the paper.}
\label{table:symbols}
\end{table}

\subsection{Monte Carlo Tree Search}

Monte Carlo Tree Search (MCTS, \citealp{MCTS_survey}) is a heuristic search algorithm renowned for its efficacy in decision-making processes within complex, stochastic environments.
Rooted in the principles of Monte Carlo simulations and tree search, MCTS has emerged as a cornerstone methodology for sequential decision problems, particularly in domains characterized by vast state spaces and incomplete information.

\begin{algorithm}[htb] 
\caption{MCTS with UCT}
\label{algo:MCTS_UCT} 
    
    \SetKw{Func}{Function:}
    \SetKw{Return}{Return:}
    
    \Func MCTS\_UCT($s_0$, $T$)
    
    \KwIn{original state $s_0$, search steps $T$}
    \KwOut{best leaf state $s*$}
    
    new root node $v_r$ of the tree\; 
     
    $v_r$.state $\leftarrow s_0$ \;
    $v_0 \leftarrow v_r$\;
	 
	 \While{\textnormal{current search steps} $< T$}{
	   $v_l \leftarrow $ tree\_policy($v_0$)\;
	   $\Delta \leftarrow $ default\_policy($v_l$.state)\;
	   backup($v_l, \Delta$)\;
	 }
	 
	 $v_{best} \leftarrow v_0$\;
	 \While{\textnormal{not $v_{best}$.terminal()} }{
	 $v_{best} \leftarrow$ best\_child($v_{best}$, 0)\;
	 }
	 
	 $s* \leftarrow$ $v_{best}$.state\;
	 
	 \Return{$s*$}
\end{algorithm}

As shown in Algorithm \ref{algo:MCTS_UCT}, it operates through four recursive phases: selection (func \textit{tree policy}), expansion (func \textit{expand}), simulation (func \textit{default policy}), and backpropagation (func \textit{backup}).
During the selection phase, the algorithm traverses the tree from the root node to a leaf node by applying a policy—commonly the Upper Confidence Bound for Trees (UCT, \cite{UCT}) —to navigate the trade-off between exploring less-visited nodes and prioritizing nodes with high reward estimates.
\begin{equation}
\begin{aligned}
\label{eq:raw_UCT}
\mathrm{S_{UCT}(v')} & = \mathrm{S_{Exploit}(v')} + \mathrm{S_{Explore}(v')} \\
&=	\frac{Q(v')}{N(v')} + C_p \sqrt{\frac{2 \ln N(v)}{N(v')}}
\end{aligned}
\end{equation}
Upon reaching a leaf node, the tree is expanded by adding one or more child nodes, thereby incrementally refining the search space.
A Monte Carlo simulation (or rollout) is then executed from the newly expanded node to estimate the potential reward of the path, typically through random or lightweight heuristic-based playouts.
Finally, the results of the simulation are backpropagated through the traversed nodes to update their statistical metrics (e.g., visit counts and cumulative rewards), refining future search iterations.

\section{SELT: Self-Evaluation Tree Search}

\subsection{LLM Task Decomposition}

LLMs have achieved significant advancements across various tasks, including text translation and domain-specific question answering (QA). While they are capable of handling a wide range of tasks, the accuracy of their responses can vary. Our preliminary experiments indicate that the phrasing of questions can influence the model's responses to the same query. To enhance the LLM's inference capabilities, we decompose complex tasks into clearer, more manageable atomic subtasks.

\begin{figure}[htb]
  \centering
  \includegraphics[width=0.95\linewidth]{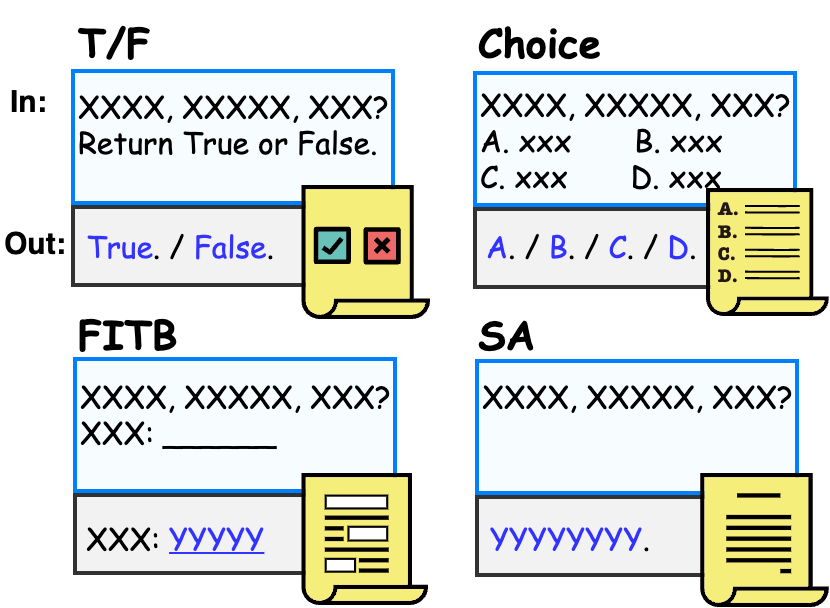}
  \caption{Four Task Type for LLMs. (T/F, Choice, FITB, SA)}
  \label{fig:task_type}
\end{figure}

We categorize tasks into the following types:

\begin{itemize}
    \item \textbf{T/F} (True-or-False Statement): Evaluate and assess information to form judgments or conclusions.
    \item \textbf{Choice}: Select the most appropriate option from a set of alternatives.
    \item \textbf{FITB} (Fill-in-the-Blank): Complete sentences or passages by inserting missing words or phrases.
    \item \textbf{SA} (Short Answer): Provide answers to specific questions based on available information.
\end{itemize}

\begin{figure}[htb]
  \centering
  \includegraphics[width=0.95\linewidth]{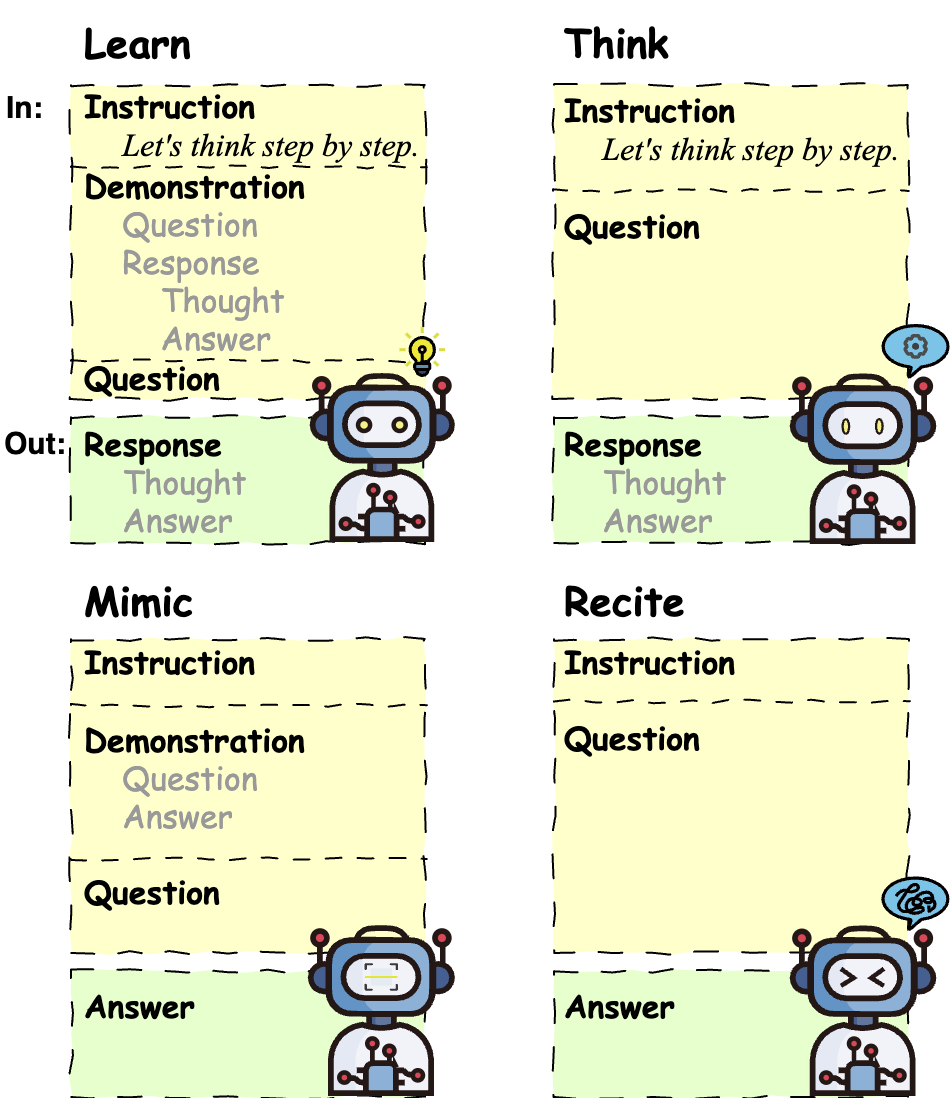}
  \caption{Four Inference Mode of LLMs. (Learn, Think, Mimic, Recite)}
  \label{fig:inference_mode}
\end{figure}

We define the following modes for inference:

\begin{itemize}
    \item \textbf{Learn}: Acquire new skills by thinking about the task and mimic the solution given by the example.
    \item \textbf{Think}: Engage in reasoning or problem-solving processes to derive conclusions.
    \item \textbf{Mimic}: Imitate or replicate the style, tone, or content of a given example.
    \item \textbf{Recite}: Reproduce information verbatim from a source or memory.
\end{itemize}

By systematically decomposing tasks into these atomic components and defining clear inference modes, we aim to improve the LLM performance and reliability across various applications.

\subsection{Node Composition in the Reasoning Tree}

\begin{figure}[htb]
  \centering
  \includegraphics[width=0.8\linewidth]{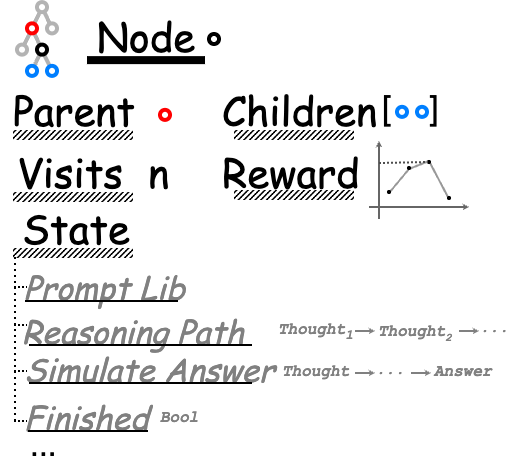}
  \caption{Node Composition. Structure info is directly stored in the node. Reasoning info is stored in the state of it.}
  \label{fig:node_state}
\end{figure}

As illustrated in Figure \ref{fig:node_state}, we design a hierarchical information storage structure within the node.
Task-specific information, such as the prompt library and reasoning paths, is stored in the node state.
When a new task is provided, we only need to modify the prompt in the state and observe the simulated response, without needing to concern ourselves with the tree search process outside the state.

\subsection{Unsupervised Semantic-Aware Clustering}
\label{sec:clustering}

In this section, we present our approach to answer clustering.
As the reasoning tree expands, the number of simulated answers at each node increases.
To effectively analyze the diverse responses of the Large Language Model (LLM) to a specific question, it is essential to cluster semantically similar responses.

\textbf{Similar Matrix Calculation}: With the simulated answers collected from each node in the search tree, we use the TF-IDF (Term Frequency – Inverse Document Frequency, \citealp{TF-IDF}) algorithm to obtain the representation vectors of each answer.
Then, the cosine similarity between these representation vectors is calculated to form a similarity matrix. 

\textbf{Clustering Number Determination}: To better aggregate different answers, we next determine the number of clusters using the similarity matrix.
The similarity matrix $G$ serves as the degree matrix of a graph and can correspond to an undirected graph.
The Laplacian matrix $L$ of graph $G$ is computed using the equation $ L = D - A $, where $L$ is the Laplacian matrix of graph $G$, $D$ is the similarity matrix (i.e., the degree matrix of graph $G$), and $A$ is the adjacency matrix of graph $G$.
Next, we compute the eigenvalues of matrix $L$ and sort them in ascending order. 
The largest difference between consecutive eigenvalues is identified, and the corresponding eigenvalue jump is taken as the number of clusters.
This jump reflects a significant transition in the graph structure from several connected components to a more complex structure.
The number of clusters is set to this eigenvalue jump number, and if the number of clusters is too large, it is capped at a fixed upper limit of 5.

\textbf{Reasoning Node Spectral Clustering}: Finally, spectral clustering \cite{SpectralClustering} is applied to classify all the reasoning nodes.
Then we obtain multiple node sets, from which the highest-scoring answer in each cluster is selected as the representative answer for that cluster.
During the scoring phase, the LLM can assess the new simulated answer with those representative answers as reference.

\begin{figure*}[htb]
  \centering
  \includegraphics[width=0.95\linewidth]{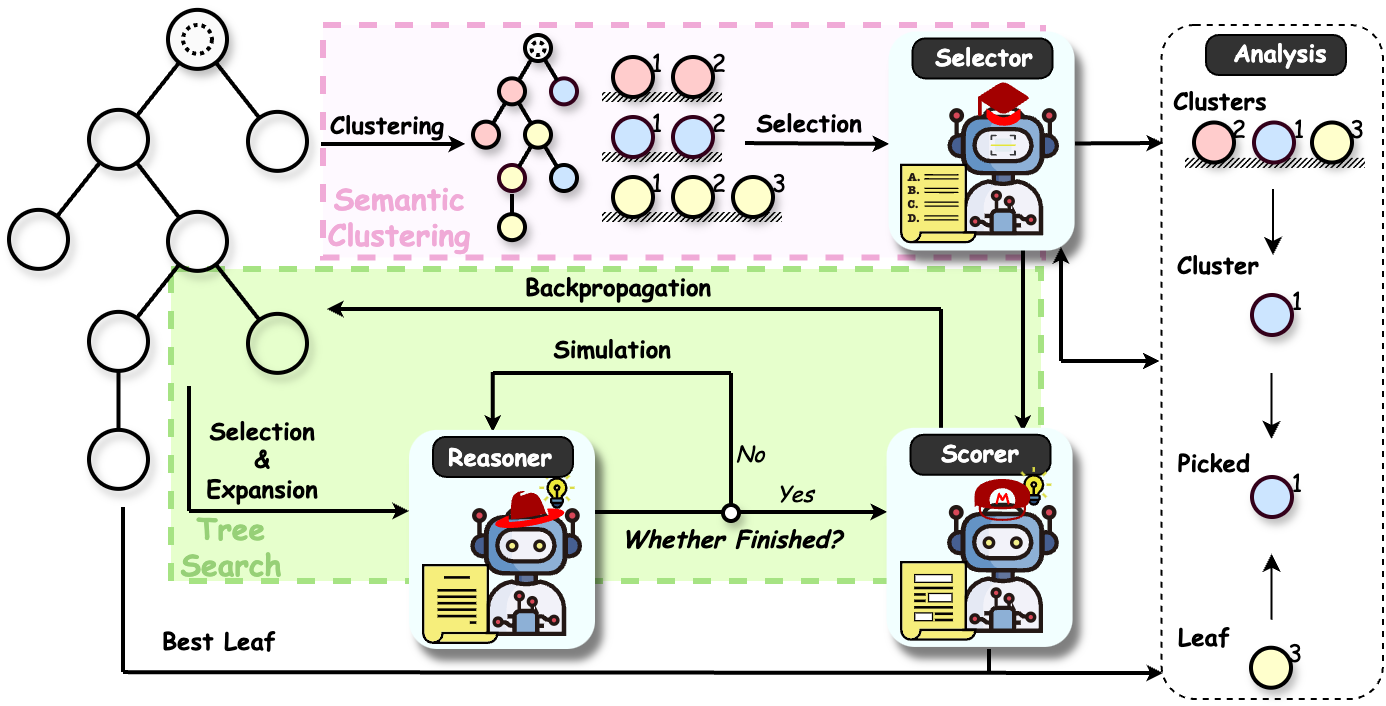}
  \caption{The Whole Procedure of SELT.}
  \label{fig:MCTS}
\end{figure*}

\subsection{Search Procedure for Better Reasoning}
\label{sec:search}



Traditional Monte Carlo Tree Search (MCTS) methods tend to exhaustively explore all possible actions at each node along the current best path.
This approach is inefficient for large model inference processes.
Additionally, current methods that apply MCTS to LLM inference require fine-tuning an extra reward model. 
Similar to fine-tuning the base model, this process lacks generality.
In this section, we propose an optimized tree search strategy for LLMs to enhance exploration effectiveness and eliminate the need for additional training.
The procedure is shown in Figure \ref{fig:MCTS}.
After preliminary evaluation, we adopt the binary tree as the search tree.
The number of nodes grows exponentially with increasing degree, providing a sufficiently large reasoning space in this setting.

\begin{algorithm}[htb] 
\caption{Selection \& Expansion}
\label{algo:selection&expansion} 
    
    \SetKw{Func}{Function:}
    \SetKw{Return}{Return:}
    
    \Func tree\_policy($v$)
    
    \KwIn{original node $v$}
    \KwOut{selected node $v$}
    
    $v_{initial} \leftarrow v$\;
    \While{\textnormal{not $v$.terminal()} }{
    \uIf{\textnormal{not $v$.children} }{\Return{\textnormal{expand($v$)}}}
    \uElseIf{\textnormal{random(0,1) < 0.5}}{$v \leftarrow$ best\_child($v, C_p$)\;}
    \uElse{
    \eIf{\textnormal{not $v$.fully\_expanded()}}{
    $v \leftarrow$ expand($v$)\;
    \eIf{\textnormal{not $v$.visits}}{\Return{$v$}}{\Return{\textnormal{tree\_policy($v_{initial}$)}}}
    }{best\_child($v, C_p$)\;}
    }
    }
    
    \Return{$v$}
\end{algorithm}

\noindent\textbf{• Selection \& Expansion}


The optimized \textit{tree\_policy} function, as shown in Algorithm \ref{algo:selection&expansion}, prioritizes expanding the depth of the tree to explore subsequent steps.
When the current node is not a leaf, it has a 50\% chance of directly proceeding to the best child node (calculated with Equation \ref{eq:exploitation} \& \ref{eq:exploration}) for the next search step, rather than attempting other untried actions at the current node.

Additionally, we have refined the formula used in the \textit{best\_child} function to calculate the Upper Confidence Bound for Trees (UCT, like in Equation \ref{algo:MCTS_UCT}) score for selecting the best child node.
As previously mentioned, the search tree is a binary tree, which means the action space in LLM tree search is binary. Let the current node be \( v \) and the child node to be evaluated be \( v' \).

The original UCT formula (Equation \ref{eq:raw_UCT}) consists of two components: exploitation and exploration.
Considering that LLMs often generate answers without a gold-standard reference, making objective evaluation challenging, we aim to enhance the confidence in evaluating the current answer by increasing the number of evaluations.
The higher the evaluation count, the more reliable the overall score distribution.
Otherwise, the node's score should tend toward the average score of the search tree.
\begin{align}
\label{eq:exploitation}
\mathrm{S_{LLM\_Exploit}}(v') &= \frac{\mu_{\mathrm{Tree}}\cdot C_\beta + Q(v')}{C_\beta + N(v')} \\
\label{eq:exploration}
\mathrm{S_{LLM\_Explore}}(v') &= C_p \sqrt{\frac{2 \ln N(v')}{N(v')}}
\end{align}
Therefore, we adopt Bayesian Averaging, modifying the exploitation component of the formula to Equation \ref{eq:exploitation}, where \( \mu_{\text{Tree}} \) denotes the average score of all simulated answers in the tree, and \( C _{\beta}\) is the expected number of evaluations (representing confidence, set here to 2).

To align with the aforementioned strategy of prioritizing deeper exploration, we also optimize the exploration component of the UCT formula like in Equation \ref{eq:exploration}.
The original exploration term is \( C_p \sqrt{\frac{2 \ln N(v)}{N(v')}} \). 
Let the visit counts of the left and right child nodes of the binary tree be \( a \) and $b$, respectively, with \( a \leq b \) and \( b = k \cdot a \), where \( k \geq 1 \).
Using the original UCT formula, the exploration scores for the two nodes are: $\sqrt{\frac{\ln (a + b)}{a}} \geq \sqrt{\frac{\ln (a + b)}{b}}$.
The optimized exploration term is \( C_p \sqrt{\frac{2 \ln N(v')}{N(v')}}, \) leading to exploration scores: $\sqrt{\frac{\ln a}{a}} \geq \sqrt{\frac{\ln b}{b}}$.
This optimization maintains the original monotonicity and results in a more gradual exploration score ratio, encouraging the prioritization of the best child node.

This refined strategy effectively balances exploration and exploitation, reduces redundant reasoning paths, and mitigates hallucination in LLM-based tree search algorithms. 

\begin{algorithm}[htb] 
\caption{Simulation}
\label{algo:simulation} 
    
    \SetKw{Func}{Function:}
    \SetKw{Return}{Return:}
    
    \Func default\_policy($s$)
    
    \KwIn{input state $s$}
    \KwOut{reward for state $r(s)$}
    
	\While{\textnormal{not $s$.terminal()}}{
	choose $a \in s$.all\_actions() uniformly at random\;
	$s \leftarrow$ next\_state$(s,a)$\;
	}
	 \Return{$r(s)$}
\end{algorithm}

\begin{algorithm}[htb] 
\caption{Backpropagation}
\label{algo:backpropagation} 
    
    \SetKw{Func}{Function:}
    \SetKw{Return}{Return:}
    
    \Func backup($v$, $\Delta$)
    
    \KwIn{input node $v$, reward score $\Delta$}
    
	\While{$v$ is no null}{
	$v$.visits $\leftarrow v$.visits $+ 1$\;
	$v$.reward $\leftarrow v$.reward $+ \Delta$\;
        $v$.reward\_list.append($\Delta$)\;
	$v \leftarrow v$.parent\;
	}
\end{algorithm}

\noindent\textbf{• Simulation}

In this step, the LLM completes the response following the established reasoning path as in Algorithm \ref{algo:simulation}. 
Before calculating the reward for the final answer, we perform semantic clustering on all nodes in the reasoning tree based on their simulated answers, as detailed in Section \ref{sec:clustering}. 
By using the representative response of each cluster as a reference criterion, the LLM can more accurately assess the new simulated answer.

\noindent\textbf{• Backpropagation}

After obtaining the reward score for the current node, we backpropagate to update the rewards along the reasoning path up to the root node like in Algorithm \ref{algo:backpropagation}.
The reward for each child node is recorded in the node information.
The reward list could be used for further scoring stability analysis.

\noindent\textbf{• Analysis}

This step is about how to obtain the best answer from the reasoning tree after search loops.
We try to get answers through two ways and find the better one between them.

\noindent\textbf{Leaf}: With UCT (set $C_p$ to 0 as in Algorithm \ref{algo:MCTS_UCT}), the answer from the best leaf node is marked as "Leaf". \newline
\noindent\textbf{Cluster}: As mentioned in Section \ref{sec:clustering}, we cluster all nodes in the tree and choose the representative node of each cluster. Finally, the best answer chosen by the LLM among these representative answers is marked as "Cluster". \newline
\noindent\textbf{Picked}: Choose 1 of the 2 between "Leaf" and "Cluster" to get the picked answer.

\begin{table*}[htb]
\centering
\resizebox{\textwidth}{!}{
\begin{tabular}{lcccccccccccccc}
\toprule
\multicolumn{1}{c}{\textbf{Method}} & \multicolumn{4}{c}{\textbf{Mathematics}}                                &           & \multicolumn{4}{c}{\textbf{Physics}}                                &           & \multicolumn{4}{c}{\textbf{Chemistry}}                            \\
\midrule
\textbf{1-shot}                     & \multicolumn{4}{c}{32.00}                                         &           & \multicolumn{4}{c}{40.20}                                           &           & \multicolumn{4}{c}{52.00}                                         \\
\textbf{1-shot CoT}                 & \multicolumn{4}{c}{49.00}                                         &           & \multicolumn{4}{c}{71.57}                                           &           & \multicolumn{4}{c}{61.00}                                         \\
\midrule
\textbf{SELT} & \textbf{S$_\mathrm{raw}$}    & \textbf{S$_\alpha$}    & \textbf{S$_\beta$}    & \textbf{S$_{\alpha + \beta}$}    & \textbf{} & \textbf{S$_\mathrm{raw}$}     & \textbf{S$_\alpha$}    & \textbf{S$_\beta$}     & \textbf{S$_{\alpha + \beta}$}    & \textbf{} & \textbf{S$_\mathrm{raw}$}    & \textbf{S$_\alpha$}    & \textbf{S$_\beta$}    & \textbf{S$_{\alpha + \beta}$}    \\
• \textbf{Sentence-Level} & & & & & & & & & & & & & & \\
\quad\textbf{Leaf} (Raw)        & 39.00          & \textbf{53.00} & \textbf{50.00} & \textbf{49.00} &           & \textbf{71.57}  & \textbf{78.43} & 67.65           & \textbf{77.45} &           & 56.00          & 54.00          & 52.00          & \textbf{63.00} \\
\quad\textbf{Cluster}                    & 48.00          & 42.00          & 45.00          & \textbf{50.00} &           & 60.78           & 70.59          & 64.71           & \textbf{74.51} &           & 52.00          & 52.00          & 41.00          & 59.00          \\
\quad\textbf{Picked}                     & \textbf{51.00} & \textbf{51.00} & \textbf{49.00} & \textbf{50.00} &           & 67.65           & \textbf{77.45} & 67.65           & \textbf{80.39} &           & 55.00          & 58.00          & 45.00          & \textbf{64.00} \\
\quad\textbf{Both}                       & \textbf{62.00} & \textbf{57.00} & \textbf{60.00} & \textbf{53.00} &           & \textbf{81.37}  & \textbf{81.37} & \textbf{80.39}  & \textbf{84.31} &           & \textbf{66.00} & \textbf{61.00} & 59.00          & \textbf{74.00} \\
\quad\textbf{Clusters}                   & \textbf{85.00} & \textbf{84.00} & \textbf{85.00} & \textbf{76.00} &           & \textbf{100.00} & \textbf{98.04} & \textbf{98.04}  & \textbf{97.06} &           & \textbf{95.00} & \textbf{92.00} & \textbf{95.00} & \textbf{90.00} \\
• \textbf{Response-Level} & & & & & & & & & & & & & & \\
\quad\textbf{Leaf} (Raw)        & 47.00          & \textbf{60.00} & 46.00          & 44.00          &           & \textbf{72.55}  & \textbf{78.43} & \textbf{76.47}  & \textbf{75.49} &           & \textbf{62.00} & \textbf{62.00} & 59.00          & 58.00          \\
\quad\textbf{Cluster}                    & 38.00          & \textbf{56.00} & 41.00          & \textbf{50.00} &           & 59.80           & \textbf{73.53} & 65.69           & 66.67          &           & 48.00          & \textbf{63.00} & 50.00          & 54.00          \\
\quad\textbf{Picked}                     & \textbf{53.00} & \textbf{63.00} & \textbf{49.00} & 46.00          &           & \textbf{72.55}  & \textbf{77.45} & \textbf{78.43}  & \textbf{76.47} &           & 60.00          & \textbf{62.00} & 59.00          & 57.00          \\
\quad\textbf{Both}                       & \textbf{56.00} & \textbf{63.00} & \textbf{53.00} & \textbf{54.00} &           & \textbf{74.51}  & \textbf{79.41} & \textbf{81.37}  & \textbf{80.39} &           & \textbf{67.00} & \textbf{65.00} & \textbf{67.00} & \textbf{64.00} \\
\quad\textbf{Clusters}                   & \textbf{88.00} & \textbf{78.00} & \textbf{86.00} & \textbf{75.00} &           & \textbf{98.04}  & \textbf{93.14} & \textbf{100.00} & \textbf{90.20} &           & \textbf{90.00} & \textbf{82.00} & \textbf{94.00} & \textbf{81.00} \\
\bottomrule
\end{tabular}
}
\caption{Experiment Results on the Benchmark MMLU. SELT-Leaf with S$_\mathrm{raw}$ can be viewed as the raw MCTS.}
\label{table:exp_MMLU}
\end{table*}

\begin{table*}[htb]
\centering
\resizebox{\textwidth}{!}{
\begin{tabular}{lcccccccccccccc}
\toprule
\multicolumn{1}{c}{\textbf{Method}} & \multicolumn{7}{c}{\textbf{Single}} & \multicolumn{7}{c}{\textbf{Multiple}} \\
\textbf{}                           & \textbf{Format}      & \multicolumn{3}{c}{\textbf{Tool}}                                  & \multicolumn{3}{c}{\textbf{Param}}                                 & \textbf{Format}      & \multicolumn{3}{c}{\textbf{Tool}}                                  & \multicolumn{3}{c}{\textbf{Param}}                                 \\
    & \textbf{}            & \textbf{P}           & \textbf{R}           & \textbf{F1}          & \textbf{P}           & \textbf{R}           & \textbf{F1}          & \textbf{}            & \textbf{P}           & \textbf{R}           & \textbf{F1}          & \textbf{P}           & \textbf{R}           & \textbf{F1}          \\
\midrule
\textbf{1-shot}                     & {\ul 100.00}         & {\ul 76.03}          & 92.00                & {\ul 83.26}          & {\ul 55.87}          & {\ul 82.14}          & {\ul 66.51}          & 99.00                & 92.33                & 90.60                & 91.46                & 74.81                & 86.58                & 80.27                \\
\textbf{1-shot CoT}                 & 99.00                & 63.33                & {\ul 95.00}          & 76.00                & 48.87                & 77.38                & 59.91                & {\ul 100.00}         & {\ul 94.36}          & {\ul 94.36}          & {\ul 94.36}          & {\ul 82.26}          & {\ul 87.12}          & {\ul 84.62}          \\
\midrule
\textbf{MCTS}                       & 95.00                & 75.63                & 90.00                & 82.19                & \textbf{59.42}       & 73.21                & 65.60                & 86.00                & \textbf{94.89}       & 81.50                & 87.69                & 82.16                & 73.35                & 77.50                \\
\textbf{SELT} & & & & & & & & & & & & & & \\
• $\mathbf{S_{\alpha}}$ & & & & & & & & & & & & & & \\
\quad\textbf{Leaf} (Raw)                      & 95.00                & \textbf{76.23}       & 93.00                & \textbf{83.78}       & \textbf{57.48}       & 73.21                & 64.40                & 87.00                & \textbf{96.28}       & 81.19                & 88.10                & 82.23                & 71.20                & 76.32                \\
\quad\textbf{Cluster}                    & 99.00                & \textbf{78.05}       & \textbf{96.00}       & \textbf{86.10}       & \textbf{58.18}       & 76.19                & 65.98                & 99.00                & \textbf{95.15}       & 92.16                & 93.63                & \textbf{82.78}       & 85.15                & 83.95                \\
\quad\textbf{Picked}                     & 99.00                & \textbf{80.67}       & \textbf{96.00}       & \textbf{87.67}       & \textbf{57.87}       & 74.40                & 65.10                & \textbf{100.00}      & \textbf{95.51}       & 93.42                & \textbf{94.45}       & \textbf{83.13}       & 86.40                & \textbf{84.74}       \\
• $\mathbf{S_{\alpha + \beta}}$ & & & & & & & & & & & & & & \\
\quad\textbf{Leaf} (Raw)                      & 96.00                & \textbf{79.13}       & 91.00                & \textbf{84.65}       & \textbf{64.50}       & 76.79                & \textbf{70.11}       & 86.00                & \textbf{94.57}       & 81.82                & 87.73                & 81.98                & 74.06                & 77.82                \\
\quad\textbf{Cluster}                    & 95.00                & \textbf{75.42}       & 89.00                & 81.65                & \textbf{59.49}       & 69.05                & 63.91                & 98.00                & \textbf{95.22}       & 93.73                & \textbf{94.47}       & \textbf{83.87}       & \textbf{88.37}       & \textbf{86.06}       \\
\quad\textbf{Picked}                     & 96.00                & \textbf{77.97}       & 92.00                & \textbf{84.40}       & \textbf{62.56}       & 72.62                & \textbf{67.22}       & 96.00                & \textbf{95.08}       & 90.91                & 92.95                & \textbf{83.84}       & 84.44                & 84.14                \\
\bottomrule
\end{tabular}
}
\caption{Experiment Results on the Benchmark Seal-Tools.}
\label{table:exp_SealTools}
\end{table*}

\section{Experiments}

\subsection{Experimental Setup}

To better present the experimental results, this section outlines the experimental settings.
We take 1-shot, 1-shot CoT and standard MCTS as baselines.
The standard MCTS can be seen in the result table as SELT with Leaf and $S_{raw}$ settings.
Relevant benchmarks, models, and other parameters are listed below.

For the foundation model, we use \textbf{Llama-3.1-8B-Instruct}\footnote{\url{https://huggingface.co/meta-llama/Llama-3.1-8B-Instruct}} \cite{llama3}, which is a relatively strong model among smaller LLMs.
To accelerate inference, we deploy the vLLM framework\footnote{\url{https://github.com/vllm-project/vllm}}.
Although fine-tuning is not required, the tree search process takes a considerable amount of time, as it is based on the search step $T$ (set to \textbf{100} in our experiments). 
After verification, 100 search steps are sufficient to explore the action space of most problems.

For Benchmarks, we use the knowledge-based QA dataset MMLU\cite{MMLU} and the tool learning dataset Seal-Tools\cite{Seal-Tools}.
We only take a part of them due to the speed limit.
For MMLU, we take abstract algebra, college physics, college chemistry splits as domain maths, physics and chemistry.

When performing step-by-step inference with LLMs, it is essential to determine the level of granularity for each step.
In \textbf{sentence-level} inference, the LLM generates the next sentence of thought at each step.
In \textbf{response-level} inference \cite{MCTSr, Llama-berry}, the LLM refines the entire response generated in the previous step.
We conduct experiments at both granularities.

Except for the analysis results mentioned in Section \ref{sec:search}, we take "Both" to record whether the gold answer is between "Leaf" and "Cluster".
We also take "Clusters" to judge whether the gold answer is among all the representative answers of each cluster.
S$_{\alpha + \beta}$ is the modified UCT mentioned in Section \ref{algo:selection&expansion}.
S$_{raw}$ is the original version of UCT, S$_{\alpha}$ is the exploitation-modified UCT (as Equation \ref{eq:exploitation}), and S$_{\beta}$ is the exploration-modified UCT (as Equation \ref{eq:exploration}).

\subsection{Main Results}

\subsubsection{Knowledge QA}

MMLU evaluates the ability of LLMs to answer domain-specific questions based on their internal knowledge.
As shown in Table \ref{table:exp_MMLU}, Sentence-Level SELT outperforms the raw MCTS (Leaf with S$_{raw}$ setting) \cite{MCTS_survey} through our newly modified UCT and answer clustering strategy.
It is also better than Response-Level SELT, demonstrating that sentence-level tree search is more effective for LLMs, as it provides sufficient reasoning space for exploration.
On the other hand, Response-Level SELT tends to prioritize exploitation, and the modified exploration strategy does not perform well.
This may suggest that it requires a more aggressive exploration of the deeper layers of the reasoning tree.

\subsubsection{Tool Learning}

Seal-Tools assesses the ability of LLMs to appropriately call external tools during inference, which is crucial for the foundation models of autonomous agents.
Like in Table \ref{table:exp_SealTools}, SELT achieves greater improvements in the single-tool-calling split, while the results for multiple-tool calling are influenced by the output format.
An interesting observation is that the baseline of 1-shot CoT performs worse than 1-shot in the single-calling split.
This may be because the single-calling data does not require step-by-step reasoning.

\subsection{Experimental Analysis}

Experiments show that SELT performs better than raw MCTS in both sentence and response levels.
Improvements in UCT exploitation and semantic clustering are beneficial across multiple settings.
Compared to the response level, sentence-level tree search can explore the action space more thoroughly and with a higher upper bound.
However, the ability of LLMs to follow formats still requires attention, as it may influence the effectiveness of long-chain reasoning.

\section{Related Works}

Researchers have tried to improve the LLM capabilities by refining its training procedure.
Setting a larger model size and using much more data is proven to be effective in this phase.
While the test time scaling law \cite{test_time_scaling} of model inference also performs well.
Many studies have confirmed that exploring the inference method of the LLM can significantly improve its performance on numerous downstream tasks.
Compared with fine-tuning method, it requires lower computational resources and can be used at any time.

Since the generative language model ChatGPT came into the public sight, many prompt-based methods have been found to guide LLMs to output better.
In-context learning (ICL, \citealp{ICL}) adds similar examples into the prompt as the demonstration; then the model can imitate their thinking style for a better answer.
Chain of Thought (CoT, \citealp{CoT}) asks the model to think about the question step by step and pay more attention to details of the question.
Tree of Thought (ToT, \citealp{ToT}) uses the searching strategies of DFS and BFS for a better thought path in the reasoning tree.
ReAct \cite{ReAct} repeats the process of reasoning and observational reflection in generation.
Although these methods have been successful in encouraging the LLM to think more, it may still struggle to solve complex problems, particularly when they require deep reasoning or the ability to explore a wide range of possible solutions.

In order to further stimulate the potential of LLMs, Monte Carlo Tree Search (MCTS) is under investigation during the reasoning process.
MCTS algorithm has gained great success in the game of go.
That proves the validity of the method.
It is used to find the best possible path in a huge multi-step action space \cite{MCTS_survey}.
LLM-MCTS \cite{LLM_MCTS} combines commonsense prior belief and MCTS together to achieve effective reasoning.
RAP \cite{RAP} reconsiders the LLM as a world model to explore strategically in the vast reasoning space through MCTS.
Someworks use MCTS to construct training data for model fine-tuning (\cite{AlphaLLM, Rest-mcts}).
LATS \cite{LATS} combines MCTS and ReAct together to observe and reflect on possible problems in the path of reasoning.

\section{Conclusion}

In this paper, we introduce SELT, a novel MCTS framework that leverages the intrinsic self-evaluation capabilities of LLMs to enhance complex reasoning without relying on external reward models.
By redefining the UCT scoring mechanism and decomposing the reasoning process into atomic subtasks with semantic clustering, SELT effectively balances exploration and exploitation while mitigating redundancy and hallucination issues.
Our extensive experiments on the MMLU and Seal-Tools benchmarks demonstrate that SELT not only achieves significant improvements in answer accuracy and robustness over traditional methods such as Chain-of-Thought prompting and standard MCTS but also maintains strong generalizability across diverse reasoning tasks without task-specific fine-tuning.
These encouraging results pave the way for future work in further integrating more adaptive self-evaluation techniques, and exploring broader applications across additional reasoning domains.
Overall, SELT represents a promising step toward more efficient, reliable, and scalable reasoning processes in large language models.

\section*{Limitations}

While SELT demonstrates promising improvements in enhancing LLM reasoning capabilities, several limitations should be noted. 

• Self-Evaluation: The framework relies on the LLM’s intrinsic self-evaluation ability to modify the UCT scoring, which may not always yield accurate assessments of intermediate reasoning quality. Errors or biases in self-evaluation could propagate through the tree search, affecting the final outcomes. 

• Computational Cost: Despite efforts to improve search efficiency, the additional computational overhead introduced by tree search remains higher compared to conventional prompting methods, which could limit real-time applications. 

• Benchmarks: Due to limited computational resources, our current experiments have been restricted to specific benchmarks (e.g., knowledge-based QA and tool learning tasks), and further research is needed to assess the framework's scalability and performance on more diverse and open-ended reasoning challenges.

Addressing these limitations in future research will be crucial for further refining the LLM inference performance, ensuring broader applicability, and advancing the progress of large language models in complex reasoning.


\bibliography{custom}

\appendix
\onecolumn 
\section{More Experiment Results}
\label{sec:appendix}

\begin{table}[htb]
\centering
\resizebox{\textwidth}{!}{
\begin{tabular}{lcccccccccccccc}
\toprule
\multicolumn{1}{c}{\textbf{Method}} & \multicolumn{7}{c}{\textbf{Single}}                                                                                                                            & \multicolumn{7}{c}{\textbf{Multiple}}                                                                                                                          \\
\textbf{}                           & \textbf{Format}      & \multicolumn{3}{c}{\textbf{Tool}}                                  & \multicolumn{3}{c}{\textbf{Param}}                                 & \textbf{Format}      & \multicolumn{3}{c}{\textbf{Tool}}                                  & \multicolumn{3}{c}{\textbf{Param}}                                 \\
                                    & \textbf{}            & \textbf{P}           & \textbf{R}           & \textbf{F1}          & \textbf{P}           & \textbf{R}           & \textbf{F1}          & \textbf{}            & \textbf{P}           & \textbf{R}           & \textbf{F1}          & \textbf{P}           & \textbf{R}           & \textbf{F1}          \\
\midrule
\textbf{1-shot}                     & {\ul 100.00}         & {\ul 76.03}          & 92.00                & {\ul 83.26}          & {\ul 55.87}          & {\ul 82.14}          & {\ul 66.51}          & 99.00                & 92.33                & 90.60                & 91.46                & 74.81                & 86.58                & 80.27                \\
\textbf{1-shot CoT}                 & 99.00                & 63.33                & {\ul 95.00}          & 76.00                & 48.87                & 77.38                & 59.91                & {\ul 100.00}         & {\ul 94.36}          & {\ul 94.36}          & {\ul 94.36}          & {\ul 82.26}          & {\ul 87.12}          & {\ul 84.62}          \\
\textbf{Sent SELT}                  &                      &                      &                      &                      &                      &                      &                      &                      &                      &                      &                      &                      &                      &                      \\
\textbf{$S_{raw}$}                          &                      &                      &                      &                      &                      &                      &                      &                      &                      &                      &                      &                      &                      &                      \\
\textbf{Picked}                     & 95.00                & \textbf{80.36}       & 90.00                & \textbf{84.91}       & \textbf{63.35}       & 72.02                & \textbf{67.41}       & 93.00                & \textbf{94.54}       & 86.83                & 90.52                & \textbf{82.27}       & 80.50                & 81.37                \\
\textbf{Leaf}                       & 95.00                & 75.63                & 90.00                & 82.19                & \textbf{59.42}       & 73.21                & 65.60                & 86.00                & \textbf{94.89}       & 81.50                & 87.69                & 82.16                & 73.35                & 77.50                \\
\textbf{Cluster}                    & 92.00                & \textbf{83.02}       & 88.00                & \textbf{85.44}       & \textbf{62.03}       & 69.05                & 65.35                & 98.00                & \textbf{94.52}       & 91.85                & 93.16                & \textbf{82.38}       & 85.33                & 83.83                \\
\textbf{$S_{\alpha}$}                          &                      &                      &                      &                      &                      &                      &                      &                      &                      &                      &                      &                      &                      &                      \\
\textbf{Picked}                     & 99.00                & \textbf{80.67}       & \textbf{96.00}       & \textbf{87.67}       & \textbf{57.87}       & 74.40                & 65.10                & \textbf{100.00}      & \textbf{95.51}       & 93.42                & \textbf{94.45}       & \textbf{83.13}       & 86.40                & \textbf{84.74}       \\
\textbf{Leaf}                       & 95.00                & \textbf{76.23}       & 93.00                & \textbf{83.78}       & \textbf{57.48}       & 73.21                & 64.40                & 87.00                & \textbf{96.28}       & 81.19                & 88.10                & 82.23                & 71.20                & 76.32                \\
\textbf{cluster}                    & 99.00                & \textbf{78.05}       & \textbf{96.00}       & \textbf{86.10}       & \textbf{58.18}       & 76.19                & 65.98                & 99.00                & \textbf{95.15}       & 92.16                & 93.63                & \textbf{82.78}       & 85.15                & 83.95                \\
\textbf{$S_{\beta}$}                         &                      &                      &                      &                      &                      &                      &                      &                      &                      &                      &                      &                      &                      &                      \\
\textbf{Picked}                     & 91.00                & \textbf{80.91}       & 89.00                & \textbf{84.76}       & \textbf{62.31}       & 73.81                & \textbf{67.57}       & 92.00                & \textbf{94.52}       & 86.52                & 90.34                & 80.66                & 79.07                & 79.86                \\
\textbf{Leaf}                       & 92.00                & \textbf{78.26}       & 90.00                & \textbf{83.72}       & \textbf{59.33}       & 73.81                & 65.78                & 79.00                & 92.34                & 71.79                & 80.78                & 80.73                & 63.69                & 71.20                \\
\textbf{Cluster}                    & 94.00                & \textbf{83.78}       & 93.00                & \textbf{88.15}       & \textbf{65.10}       & 74.40                & \textbf{69.44}       & 98.00                & 94.25                & 92.48                & 93.35                & 80.75                & 84.79                & 82.72                \\
\textbf{$S_{\alpha + \beta}$}                         &                      &                      &                      &                      &                      &                      &                      &                      &                      &                      &                      &                      &                      &                      \\
\textbf{Picked}                     & 96.00                & \textbf{77.97}       & 92.00                & \textbf{84.40}       & \textbf{62.56}       & 72.62                & \textbf{67.22}       & 96.00                & \textbf{95.08}       & 90.91                & 92.95                & \textbf{83.84}       & 84.44                & 84.14                \\
\textbf{Leaf}                       & 96.00                & \textbf{79.13}       & 91.00                & \textbf{84.65}       & \textbf{64.50}       & 76.79                & \textbf{70.11}       & 86.00                & \textbf{94.57}       & 81.82                & 87.73                & 81.98                & 74.06                & 77.82                \\
\textbf{Cluster}                    & 95.00                & \textbf{75.42}       & 89.00                & 81.65                & \textbf{59.49}       & 69.05                & 63.91                & 98.00                & \textbf{95.22}       & 93.73                & \textbf{94.47}       & \textbf{83.87}       & \textbf{88.37}       & \textbf{86.06}       \\
\textbf{Resp SELT}                       & \multicolumn{1}{l}{} & \multicolumn{1}{l}{} & \multicolumn{1}{l}{} & \multicolumn{1}{l}{} & \multicolumn{1}{l}{} & \multicolumn{1}{l}{} & \multicolumn{1}{l}{} & \multicolumn{1}{l}{} & \multicolumn{1}{l}{} & \multicolumn{1}{l}{} & \multicolumn{1}{l}{} & \multicolumn{1}{l}{} & \multicolumn{1}{l}{} & \multicolumn{1}{l}{} \\
\textbf{$S_{raw}$}                          & \multicolumn{1}{l}{} & \multicolumn{1}{l}{} & \multicolumn{1}{l}{} & \multicolumn{1}{l}{} & \multicolumn{1}{l}{} & \multicolumn{1}{l}{} & \multicolumn{1}{l}{} & \multicolumn{1}{l}{} & \multicolumn{1}{l}{} & \multicolumn{1}{l}{} & \multicolumn{1}{l}{} & \multicolumn{1}{l}{} & \multicolumn{1}{l}{} & \multicolumn{1}{l}{} \\
\textbf{Picked}                     & 98.00                & 69.34                & \textbf{95.00}       & 80.17                & \textbf{56.22}       & 77.98                & 65.34                & 98.00                & 92.21                & 89.03                & 90.59                & 82.17                & 82.47                & 82.32                \\
\textbf{Leaf}                       & 98.00                & 69.06                & \textbf{96.00}       & 80.33                & 53.97                & 76.79                & 63.39                & 94.00                & 91.95                & 85.89                & 88.82                & 80.93                & 77.46                & 79.16                \\
\textbf{Cluster}                    & 97.00                & 68.89                & 93.00                & 79.15                & \textbf{58.48}       & 77.98                & \textbf{66.84}       & 99.00                & 93.97                & 92.79                & 93.38                & \textbf{85.14}       & \textbf{87.12}       & \textbf{86.12}       \\
\textbf{$S_{\alpha}$}                          &                      &                      &                      &                      &                      &                      &                      &                      &                      &                      &                      &                      &                      &                      \\
\textbf{Picked}                     & 99.00                & 62.18                & \textbf{97.00}       & 75.78                & 50.57                & 79.17                & 61.72                & 97.00                & 92.95                & 90.91                & 91.92                & \textbf{82.86}       & 83.01                & 82.93                \\
\textbf{Leaf}                       & 99.00                & 61.69                & \textbf{95.00}       & 74.80                & 50.39                & 77.38                & 61.03                & 95.00                & 93.14                & 89.34                & 91.20                & \textbf{82.57}       & 80.50                & 81.52                \\
\textbf{Cluster}                    & \textbf{100.00}      & 68.06                & \textbf{98.00}       & 80.33                & 53.60                & 79.76                & 64.11                & 99.00                & 90.55                & 93.10                & 91.81                & 80.94                & 85.87                & 83.33                \\
\textbf{$S_{\beta}$}                         &                      &                      &                      &                      &                      &                      &                      &                      &                      &                      &                      &                      &                      &                      \\
\textbf{Picked}                     & 96.00                & 74.19                & 92.00                & 82.14                & \textbf{60.95}       & 76.19                & \textbf{67.72}       & 99.00                & 93.71                & 93.42                & 93.56                & 81.23                & \textbf{87.48}       & 84.24                \\
\textbf{Leaf}                       & 96.00                & 71.54                & 93.00                & 80.87                & \textbf{56.19}       & 75.60                & 64.47                & 87.00                & 92.83                & 81.19                & 86.62                & 81.00                & 72.45                & 76.49                \\
\textbf{Cluster}                    & 97.00                & \textbf{76.42}       & 94.00                & \textbf{84.30}       & \textbf{64.90}       & 80.36                & \textbf{71.81}       & \textbf{100.00}      & 90.68                & 91.54                & 91.11                & 79.83                & 85.69                & 82.66                \\
\textbf{$S_{\alpha + \beta}$}                         &                      &                      &                      &                      &                      &                      &                      &                      &                      &                      &                      &                      &                      &                      \\
\textbf{Picked}                     & 98.00                & 64.79                & 92.00                & 76.03                & 52.07                & 75.00                & 61.46                & 97.00                & \textbf{94.48}       & 91.22                & 92.82                & 81.74                & 84.08                & 82.89                \\
\textbf{Leaf}                       & 98.00                & 65.96                & 93.00                & 77.18                & 53.25                & 77.98                & 63.29                & 97.00                & 93.85                & 90.91                & 92.36                & 80.83                & 83.72                & 82.25                \\
\textbf{Cluster}                    & 98.00                & 65.28                & 94.00                & 77.05                & 54.07                & 79.17                & 64.25                & \textbf{100.00}      & 94.06                & \textbf{94.36}       & 94.21                & 82.03                & \textbf{88.19}       & \textbf{85.00}      \\
\bottomrule
\end{tabular}
}
\caption{Full Experiment Results on the Benchmark Seal-Tools.}
\label{table:exp_SealTools_all}
\end{table}

\end{document}